\documentclass[runningheads]{llncs}
\usepackage[T1]{fontenc}
\usepackage{graphicx}
\usepackage{algorithm2e}
\usepackage{subfigure}
\usepackage{cleveref}
\crefname{appendix}{App.\negthinspace\,}{App.\negthinspace\,}
\crefname{chapter}{Chap.\negthinspace\,}{Chap.\negthinspace\,}
\crefname{equation}{Eq.\negthinspace\,}{Eq.\negthinspace\,}
\crefname{algorithm}{Alg.\negthinspace\,}{Alg.\negthinspace\,}
\crefname{section}{Sec.\negthinspace\,}{Sec.\negthinspace\,}
\crefname{subsection}{Sec.\negthinspace\,}{Sec.\negthinspace\,}
\crefname{subsubsection}{Sec.\negthinspace\,}{Sec.\negthinspace\,}
\crefname{figure}{Fig.\negthinspace\,}{Fig.\negthinspace\,}
\crefname{table}{Tab.\negthinspace\,}{Tab.\negthinspace\,}
\crefname{subfigure}{Fig.\negthinspace\,}{Fig.\negthinspace\,}
\crefname{subsubfigure}{Fig.\negthinspace\,}{Fig.\negthinspace\,}
\crefname{lstlisting}{Lst.\negthinspace\,}{Lst.\negthinspace\,}

\newcommand{\etal}{\textit{et al.\,}}

\newcommand{\ie}{\textit{i.e.}}

\usepackage{color} 

\begin{document}
\title{Unsupervised Learning for Feature Extraction and Temporal Alignment of 3D+t Point Clouds of Zebrafish Embryos}
%
\titlerunning{Temporal Alignment of 3D+t Point Clouds}
%


\author{Zhu Chen$^*$ \and
Ina Laube\thanks{Equal contrib.; funded by the German Research Foundation DFG (STE2802/1-1).} \and
Johannes Stegmaier}
\authorrunning{Chen \etal}
%
\institute{Institute of Imaging and Computer Vision, RWTH Aachen University, Germany\\
\email{\{Zhu.Chen, Ina.Laube, Johannes.Stegmaier\}@lfb.rwth-aachen.de}}


%
%

%


\maketitle              
\begin{abstract}
Zebrafish are widely used in biomedical research and developmental stages of their embryos often need to be synchronized for further analysis. We present an unsupervised approach to extract descriptive features from 3D+t point clouds of zebrafish embryos and subsequently use those features to temporally align corresponding developmental stages. An autoencoder architecture is proposed to learn a descriptive representation of the point clouds and we designed a deep regression network for their temporal alignment. We achieve a high alignment accuracy with an average mismatch of only 3.83 minutes over an experimental duration of 5.3 hours. As a fully-unsupervised approach, there is no manual labeling effort required and unlike manual analyses the method easily scales. Besides, the alignment without human annotation of the data also avoids any influence caused by subjective bias.

\keywords{Embryo Development \and Point Clouds  \and Unsupervised Learning \and Autoencoder.}
\end{abstract}
\section{Introduction}
Zebrafish are widely used model organisms in many experiments due to their fully-sequenced genome, easy genetic manipulation, high fecundity, external fertilization, rapid development and nearly transparent embryos \cite{10.1093/af/vfz020}. The spatiotemporal resolution of modern light-sheet microscopes allows imaging the embryonic development at the single-cell level \cite{Kobitski2015}. Fluorescently labeled nuclei can be detected, segmented and tracked in these data sets and the extracted 3D+t point clouds can be used for analyzing the development of a single embryo in unprecedented detail \cite{Kobitski2015}. In many experiments, an important task is to compare the level of growth between different individuals, especially the research on mutants and the growth under particular conditions like pollution and exposure to potentially harmful chemicals \cite{6867986,CastroGonzlez2014}. Thus, automatically obtaining an accurate temporal alignment that synchronizes the developmental stages of two or more individuals is an important component of such comparative analyses.

Existing approaches are mostly based on the automatic alignment of manually identified landmarks or operate in the image domain. However, to the best of our knowledge there is no approach available yet to obtain a spatiotemporal alignment of 3D+t point clouds of biological specimens in an automatic and unsupervised fashion.
In \cite{6867986}, the authors generate sets of landmarks based on the deformation of embryos, and the landmarks are paired to generate the temporal registration between two sequences of embryos. Michelin \etal \cite{7493464} formulate the temporal alignment problem as an image-to-sequence registration applicable to more complex organisms like the floral meristem. The method introduced in \cite{7163872} pairs the 3D images of ascidian embryos by finding the symmetry plane and by computing the transformation that optimizes the cell-to-cell mapping. In \cite{MCDOLE2018859}, landmarks of multiple mouse embryos are manually identified and subsequently used to automatically obtain a spatiotemporal alignment with their custom-made Tardis method. 
In the past couple of years, deep neural networks emerged as a powerful approach to learn descriptive representations of images and point clouds that can be flexibly used for various tasks. The authors of \cite{cnn_ta} use an image-based convolutional neural network and a PointNet-based \cite{pointnet} architecture in a supervised fashion to obtain an automatic staging of zebrafish embryos.

In this work, we present a deep learning-based method for the temporal alignment of 3D+t point clouds of developing embryos. Firstly, an autoencoder is employed to extract descriptive features from the point clouds of each time frame. As an autoencoder designed explicitly for point clouds, FoldingNet \cite{foldingnet} is used as the basic architecture and we propose several modifications to improve its applicability to point cloud data of developing organisms. As the next step, the extracted latent features of the autoencoder are used in a regression network to temporally align different embryos. The final output are pairwisely aligned time frames of two different 3D+t point clouds.
We show that the autoencoder learns discriminative and descriptive feature vectors that allow to recover the temporal ordering of different time frames. In addition to quantitatively assessing the alignment accuracy of the regression network, we demonstrate the effectiveness of the latent features by visualizing the reconstructed 3D point clouds and low-dimensional representations obtained with t-SNE \cite{tsne}. Being a fully-automatic and unsupervised approach, our method does not require any time-consuming human labeling effort and additionally avoids any subjective biases.

\section{Methods}
Our method is based on the FoldingNet \cite{foldingnet} as the point cloud feature extractor. Several modifications are added to both the network and the loss function. The temporal alignment is realized with a regression network using the latent features and a subsequent consistency-check is used as a postprocessing to further improve the results. Finally, we introduce a new method to synthesize validation data sets with known ground truth from a set of unlabeled embryo point clouds.
\subsection{Autoencoder}
\textbf{FoldingNet}: FoldingNet is an autoencoder specifically designed for 3D point clouds \cite{foldingnet}. In the encoder, a k-nearest-neighbor graph is built and the local information of each sub-region is extracted using the graph layers. A global max-pooling operation is applied to the local features to obtain the one-dimensional latent feature vector in the bottleneck layer. As a symmetric function, the pooling operation adds permutation invariance to cope with the disorderliness of point clouds. In the decoder, the feature vector is duplicated and concatenated with a fixed grid of points. The latent features lead to the deformation of the point grid in a 3-layer multi-layer perceptron (MLP) and a 3D structure is constructed. With a second folding operation, more details are rebuilt and the input is reconstructed. Unlike the original FoldingNet \cite{foldingnet}, we use a spherical template ($M$ evenly distributed 3D points on a spherical surface) instead of a planar template, which provides a better initialization for reconstructing spherical objects from the learned representations. Since the shape of the embryos varies from a hemisphere to an ellipsoid during development, the spherical template simplifies the folding operation and improves the reconstruction quality in combination with the Modified Chamfer Distance loss (see next section).\\

\begin{figure}
\hspace{0.38cm}
{%
\subfigure[Time frame index = 100]{%
\includegraphics[width=0.45\textwidth]{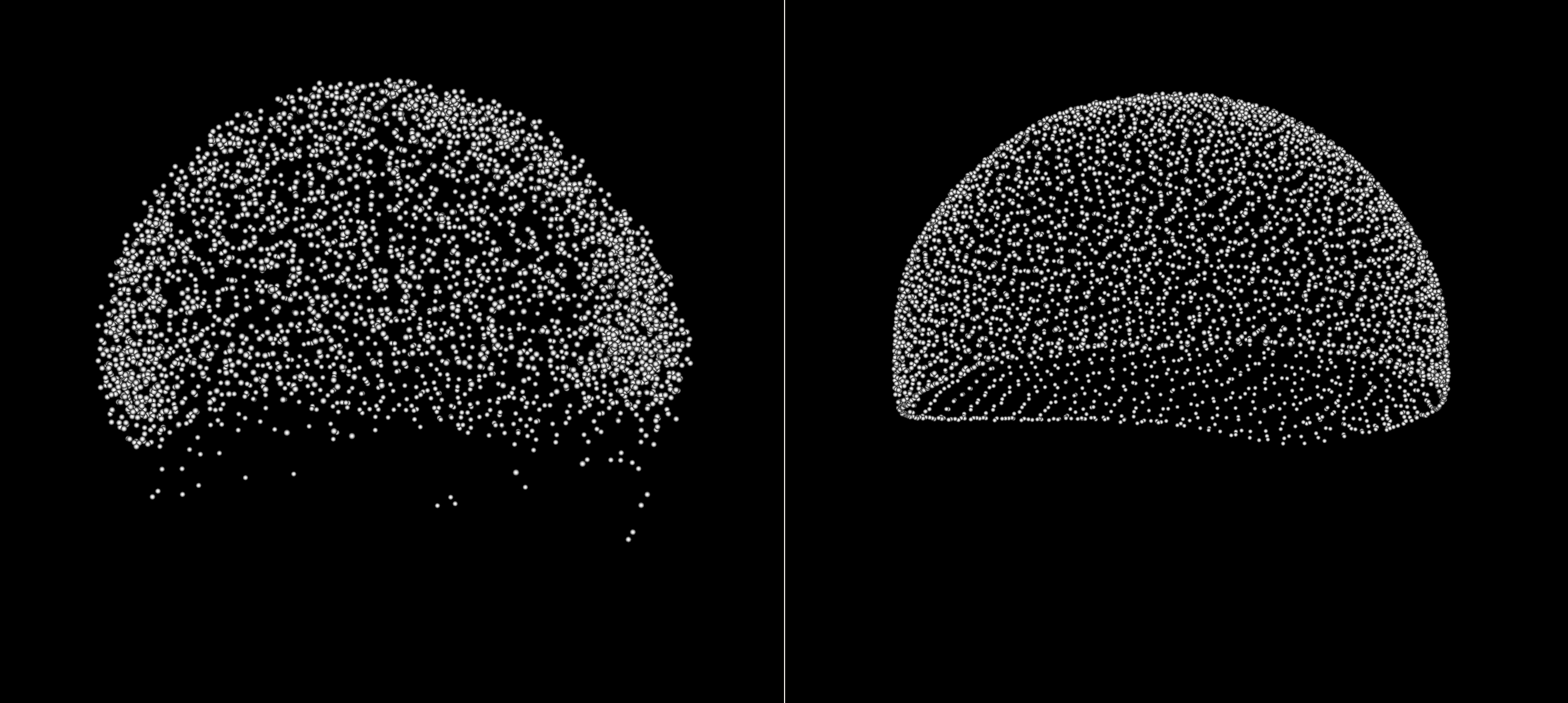}
}
\subfigure[Time frame index = 200]{%
\includegraphics[width=0.45\textwidth]{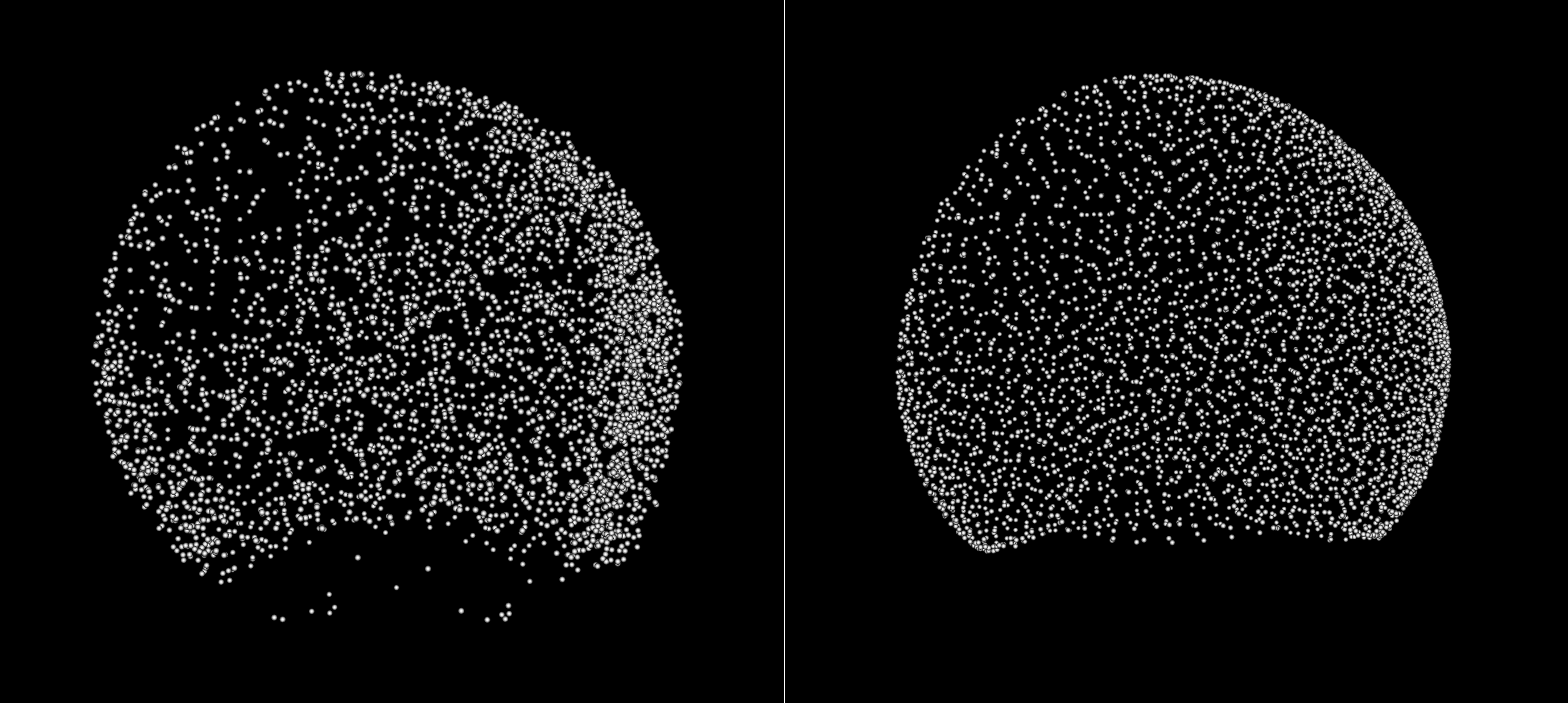}
}
}
\caption{Comparison of raw 3D point clouds (left sub-panels) and the reconstructions of our modified FoldingNet that was trained with MCD and the spherical point template (right sub-panels). Shape and density distribution of the reconstructions are nicely preserved, \ie, the learned representation successfully condenses the properties of the input point clouds (see Suppl.~\cref{fig:draw_rec} for additional examples).}
\label{fig:draw_rec_main}
\end{figure}

\noindent\textbf{Modified Chamfer Distance}: The Chamfer Distance (CD) is one of the most widely used similarity measures for point clouds and is used as the loss function of FoldingNet. The discrepancy between two point clouds is calculated as the sum of the distances between the closest pairs of points as follows:
\begin{equation}\label{eq:cd}
L_{CD}\left(S_{in}, S_{out}\right)=\frac{1}{\left|S_{in}\right|} \sum_{p \in S_{in}} \min _{q \in S_{out}}\|p-q\|_{2}+\frac{1}{\left|S_{out}\right|} \sum_{q \in S_{out}} \min _{p \in S_{in}}\|q-p\|_{2}
\textrm{,}
\end{equation}
where $S_{in}$, $S_{out}$ are the input and reconstructed point clouds, respectively. However, this approach does not consider the local density distribution since all points are treated independently. In this work, we introduce the Modified Chamfer Distance (MCD), in which the point-to-region distance replaces the point-to-point distance. The new loss is the summation of the k-nearest-neighbors of each point in the target point cloud and defined as:
\begin{eqnarray}
L_{MCD}\left(S_{in}, S_{out}\right) & = & \frac{1}{|S_{in}|}\sum_{p \in S_{in}}\frac{1}{k}\sum_{i}^{k} \min _{q_i \in S_{out}}d(p,q_i)+ \nonumber \\
 & & \frac{1}{|S_{out}|} \sum_{q \in S_{out}} \frac{1}{k}\sum_{j}^{k}\min _{p_j \in S_{in}}d(q,p_j)
\textrm{,}
\end{eqnarray}
where $d$ is the Euclidean distance between a point $p$ and its respective nearest neighbor $q_i$. As the embryos grow, the density distribution changes significantly in different parts. The utilized data set \cite{Kobitski2015} is spatially prealigned such that the animal and vegetal pole align with the y-axis and the prospective dorsal part with the positive x-axis. During epiboly, the embryo grows from a hemisphere to a complete sphere in the negative y-direction. As development progresses into the bud stage, density increases in the direction of the positive x-axis and thus the center of gravity moves to the right. We found that a FoldingNet trained with MCD consistently yielded more accurate reconstructions and alignment results compared to CD (Suppl.~\cref{fig:mcd}, Suppl.~\cref{fig:draw_rot_norm_cd}). In \cref{fig:draw_rec_main}, the input and the reconstructed point clouds of an embryo from the hold-out test set are visualized using ParaView \cite{paraview} to qualitatively illustrate the effectiveness of the FoldingNet that was trained with the MCD loss. 

\subsection{Regression Network}
\begin{figure}
\includegraphics[width=\textwidth]{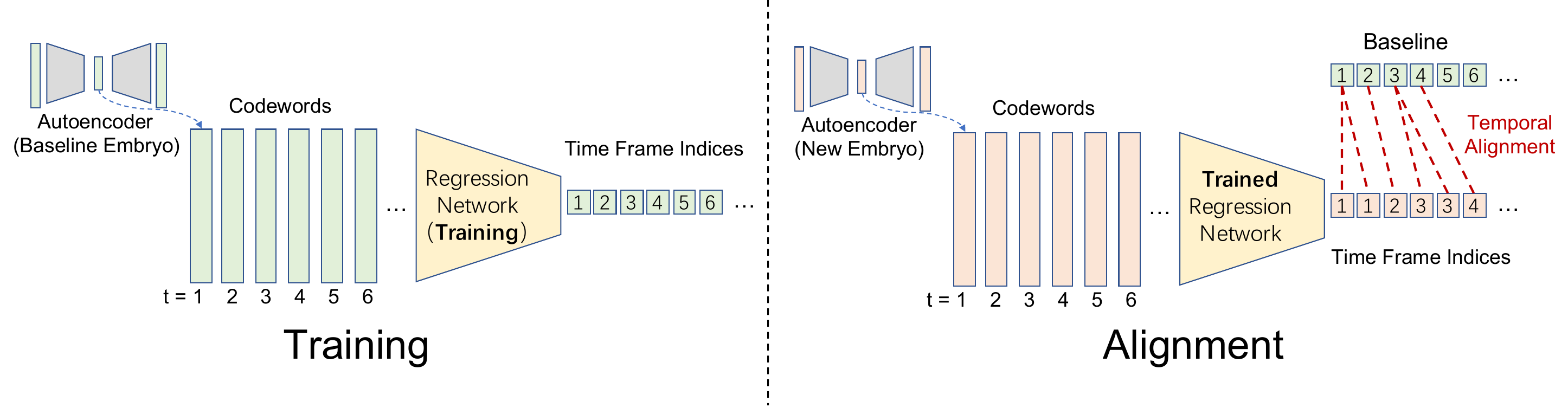}
\caption{The temporal alignment network is trained with a baseline embryo and learns to map each feature vector to the corresponding reference frame index (left). The trained network is then applied to feature vectors of a new embryo and the predicted frame index sequence indicates how the embryo should be aligned to the reference (right).} \label{fig:approach}
\end{figure}

The features extracted by the autoencoder are used to generate the temporal alignment of different embryos. We select one embryo as the reference and train an MLP regression network that maps autoencoder-generated feature vectors to frame numbers. The regression MLP consists of a sequence of fully-connected layers with ReLU activation. The input is a latent feature vector with length 256. In each dense layer, the size of the vector is reduced by a factor of two. The penultimate layer converts the 8-dimensional vector directly to the output node. The mean squared error is utilized as the loss function to compare the output to the ground truth: the sequence of time frame indices from 1 to 370. To temporally synchronize a new embryo with the reference, we present extracted features of all its frames to the trained regression network and generate a new index sequence. The desired alignment result is obtained by comparing the generated sequence with the reference embryo (\cref{fig:approach}).\\

\noindent\textbf{Postprocessing}: By definition, the sequence of time frame indices must be monotonically increasing. However, since the alignment results are generated from the point cloud of each time frame individually, the relationship between the time points in the sequence is not considered. So the generated sequences are not guaranteed to be monotonically increasing. We use a simple postprocessing strategy to generate monotonically increasing and more accurate alignment results. For an aligned sequence with oscillations, we generate its monotonically increasing upper and lower boundaries, and the desired result is obtained by calculating the mean values of those boundaries (Suppl.~\cref{fig:postprocessing}).

\section{Experimental Results and Discussions}

\subsection{Data Sets and Evaluation}
The data set used in this work was published in \cite{Kobitski2015} and consists of four wild-type zebrafish embryos that were imaged from $4.7$ to $10.0$ hpf (hours post fertilization) with one minute time intervals. Each embryo is represented as a 3D point cloud and has 370 time frames. The staging of these data sets was performed at a single time point (10 hpf) and the 370 preceding frames were selected irrespective of potential developmental differences. We thus only know that the temporal windows of the four embryos largely overlap but do not have frame-accurate annotations of the actual developmental time (see \cite{cnn_ta,Kobitski2015,Schott2018} for details).
To simplify the training approach, a fixed number of points is randomly chosen from each point cloud using the PyTorch Geometric library \cite{pyg} and we use the data loader implemented in the PyTorch 3D Points library \cite{tp3d}. We choose 4096 points since the size of the original point clouds ranges from 4160 to 19794. For improved generalizability and orientation invariance of the FoldingNet-based autoencoder, the point clouds are randomly rotated between 0 and 360 degrees along each axis as data augmentation. For additional augmentation, we generated randomized synthetic variants of the four embryos as described in \cite{cnn_ta}. Since there are no labeled data to evaluate the temporal alignment results, we introduce a new method to artificially generate ground truth for validation by randomly varying the speed of development of a selected embryo (sin, cos, Gaussian, and linear-based stretching/compression of the time axis with interpolated/skipped intermediate frames). A Gaussian-distributed point jitter ($\mu=0, \sigma^2 = 5$) is applied to the shifted embryos to make them substantially differ from the originals.

\subsection{Experimental Settings}
The FoldingNet-based autoencoder is implemented with PyTorch Lightning. The number of neighbors is set to $16$ in the KNN-graph layers and to $20$ in the MCD. The autoencoder is trained with an initial learning rate of $0.0001$ for $250$ epochs and a batch size of $1$. At each iteration, the embryo from a single time frame with the size $4096 \times 3$ is given to the network and the encoder converts it to a 1D codeword of length 256, which is the latent feature vector. For network training with the four wild-type embryos, we use a 4-fold cross-validation scheme with three embryos for training and one for testing in each fold. The regression network is trained for $700$ epochs with a learning rate of $0.00001$. The hyperparameters of the regression network are determined empirically based on the convergence of the training loss, since there is no validation or test set available. The temporal alignment is validated with the embryo from the test set to make the result independent from training the autoencoder and the embryo is aligned to its shifted variants. To reduce the influence of randomness, we repeat the alignment of each test embryo three times and take the average value of all experiments of the four embryos.
\subsection{Experimental Results}

\begin{figure}
{%
\subfigure[Cos]{%
\label{fig:cos}
\includegraphics[width=0.234\textwidth]{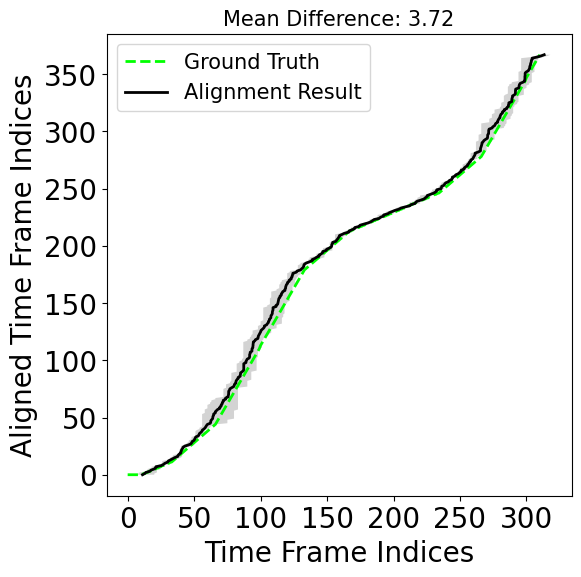}
}
\subfigure[Sin]{%
\label{fig:sin}
\includegraphics[width=0.234\textwidth]{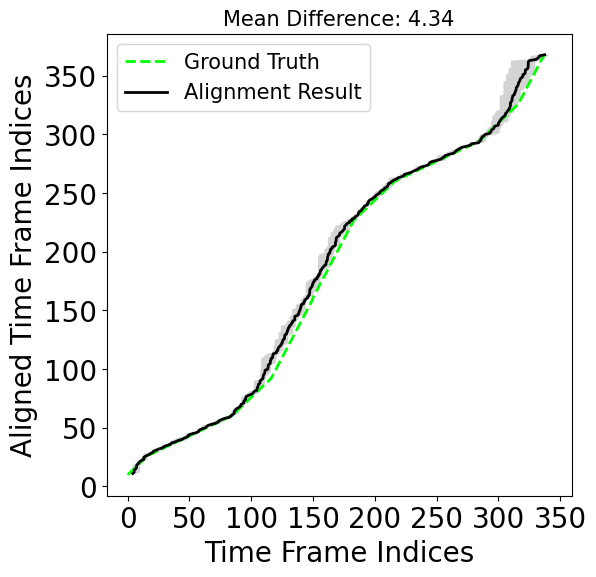}
}
\subfigure[Gaussian]{%
\label{fig:gaus}
\includegraphics[width=0.234\textwidth]{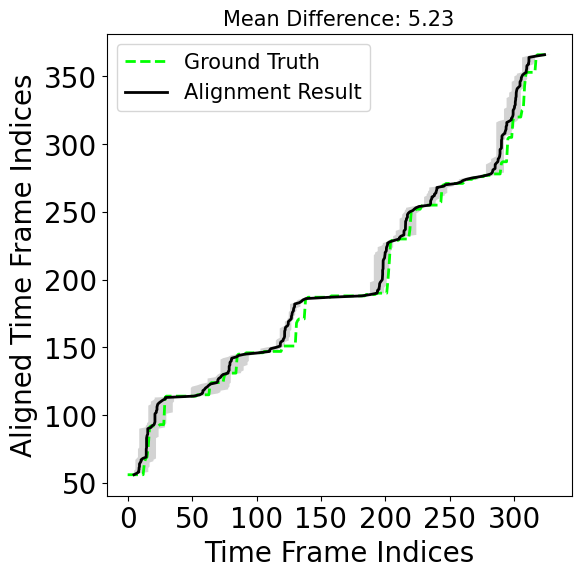}
}
\subfigure[Faster]{%
\label{fig:const}
\includegraphics[width=0.234\textwidth]{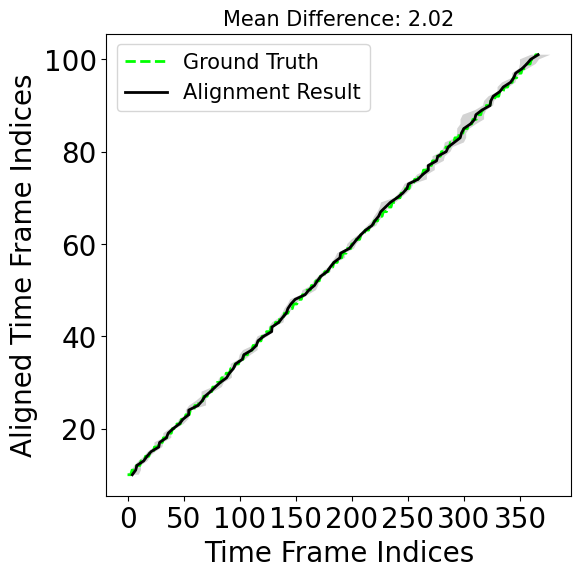}
}
}
\caption{Alignment results of the embryos with different shifting methods. The black line indicates the average result of all experiments and the area shaded in gray represents the variance. The alignment error is calculated as the average number of mismatched time frame indices.} \label{fig:draw}
\end{figure}

\noindent\textbf{Temporal Alignment Results:} The resulting alignment with different shifting methods is depicted in \cref{fig:draw}.
In \cref{fig:cos} and \cref{fig:sin}, a cosine- and sine-shifted embryo is aligned to the original one, where the cosine-based temporal shifting lets the embryo develop faster at the beginning and then slow down while the sine-based shifting is defined contrarily. In \cref{fig:gaus}, a Gaussian-distributed random difference is added to the shifted embryo. Furthermore, \cref{fig:const} illustrates an embryo that develops approximately three times faster than the original one. The alignment error is defined as the average number of mismatched indices, which is the difference between the sequence of the aligned indices and the ground truth in the x-direction. As a result, an average mismatch of only 3.83 minutes in a total developmental period of 5.3 hours is achieved (\cref{Tab:tab}).

\noindent\textbf{Impact of Spatial Transformations:} In the previous experiments, the embryos to be aligned were generated by changing the development speed and by scattering the points to make them different from the original ones. However, the embryos to be aligned in real applications could be located and oriented differently in space. Thus, we add some data augmentation for the alignment network to increase its rotation invariance. The point clouds are randomly rotated between $\pm$20 degrees along each axis. Moreover, the shifted embryos to be aligned are rotated between $\pm$15 degrees. 
The average alignment error obtained for rotated embryos is 3.48 minutes (Suppl.~\cref{fig:draw_rot}). However, a larger variance can be observed, which indicates that rotation can have an impact on the alignment accuracy in some cases. Since the overall accuracy is still high, however, our approach proofs to be robust for aligning embryos with slightly varying orientations.
In addition to the orientation, embryos may also be positioned differently in the sample chamber. To potentially improve the translation invariance, we tested if centering all point clouds at the origin of the coordinate axes before inputting them to the alignment network has a positive effect (Suppl.~\cref{fig:draw_rot_norm}). Although centering makes the approach invariant to spatial translation, we find that the alignment accuracy is reduced (aligned sequences have a more significant variance, and the average error increases to $5.74$ minutes). We hypothesize that the relative displacement of the point cloud's centroid to the origin of the coordinate system (which is removed by centering) may play an important role in determining the developmental stage and the level of completion of the epiboly phase. An overview of obtained alignment results is provided in \cref{Tab:tab}.
\begin{table}
\centering
\caption{Results of the average temporal alignment errors in minutes.}\label{Tab:tab}
\begin{tabular}{|l|l|l|l|l|l|}
\hline
\bfseries Shifting Method & \bfseries Cosine & \bfseries Sine & \bfseries Gaussian & \bfseries Faster & \bfseries Average\\
\hline
Scattering & 3.72 & 4.34 & 5.23 & 2.02 & 3.83\\
Scattering + Rot. & 3.90 & 2.89 & 5.00 & 2.13 & 3.48\\
Scattering + Rot. + Cent. & 7.16 & 5.64 & 7.38 & 2.78 & 5.74\\
\hline
\end{tabular}
\end{table}

\begin{figure}
\centering
{
\subfigure[Original]{%
\label{fig:tsneo}
\includegraphics[width=0.25\textwidth]{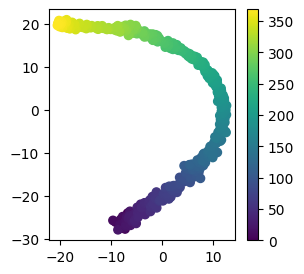}
}
\subfigure[Rotated]{%
\label{fig:tsner}
\includegraphics[width=0.25\textwidth]{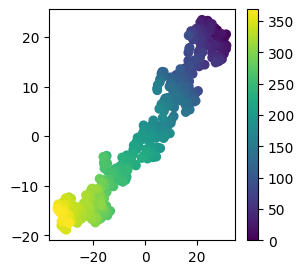}
}
\subfigure[Rotated and origin-centered]{%
\label{fig:tsnecr}
\includegraphics[width=0.25\textwidth]{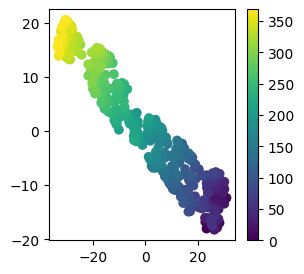}
}
}
\caption{Visualization of the feature vectors using the t-SNE algorithm. The color-code represents the time frame index and changes smoothly as the time increases.}\label{fig:tsne}
\end{figure}

\noindent\textbf{Visualizing the Learned Representation:}
To confirm that the autoencoders actually learned a representation suitable for temporal alignment, we visualize a chronologically color-coded scatter plot of the learned 256-dimensional feature vectors of all 370 time frames using the t-SNE algorithm \cite{tsne} (\cref{fig:tsne}).
The original features are clustered to a narrow band and the color changes smoothly as the index increases, which indicates that different time frames are well distinguishable using the representation learned by the autoencoder. When the point clouds are rotated and centered, the projections of the feature vectors become more dispersed as illustrated in \cref{fig:tsner} and \ref{fig:tsnecr}. Nevertheless, the color changes smoothly and the features are suitable for distinguishing different time frames. This is in line with the observed increased variance for the temporal alignment results with maintained good average alignment accuracy.

\section{Conclusion}
In this work we present a fully-unsupervised approach to temporally align 3D+t point clouds of zebrafish embryos. A FoldingNet-based autoencoder is implemented to extract low-dimensional descriptive features from large-scaled 3D+t point clouds of embryonic development. Several modifications are made to the network and the loss function to improve their applicability for this application. The embryos are temporally aligned by applying a regression network to the features extracted by the autoencoder. A postprocessing method is designed to provide consistent and accurate alignments. As no frame-accurate ground truth is available yet, we assess the effectiveness of our method via a 4-fold cross validation and a synthetically generated ground truth. An average mismatch of only 3.83 minutes in a developmental period of 5.3 hours is achieved. Finally, we performed several ablation studies to show the impact of rotation and spatial translation of the point clouds to the alignment results. By aligning embryos with different spatial locations and deflected central axis, a relatively small error rate of 5.74 minutes can still be achieved. According to feedback from a biological expert the achievable manual alignment accuracy is on the order of $30$ minutes and potentially exhibits intra- and inter-rater variabilities. As the first unsupervised method designed for the automatic spatiotemporal alignment of 3D+t point clouds, our method achieves high accuracy and eradicates the need for any human interaction. This will particularly help to minimize human effort, to speedup experimental analysis and to avoid any subjective biases.

In future works, our approach could be applied to more data sets and other model organisms with different scales and development periods to further validate its applicability. We’re currently conducting an extensive effort to obtain frame-accurate manual labels from multiple raters in a randomized study, to better assess the actual performance that we can expect under real-world conditions including intra- and inter-rater variability. In the long term, we envision an iteratively optimized spatiotemporal average model of multiple wild-type embryos to finally obtain a 3D+t reference atlas that can be used to precisely analyze developmental differences of corresponding anatomical regions across experiments.

\renewcommand{\figurename}{Suppl.~Fig.}
\renewcommand{\algorithmcfname}{Suppl.~Alg.}
\setcounter{figure}{0}

\newpage
\section*{Supplementary Material}

\begin{figure}[htb!]
\centering
{%
\subfigure[x-axis: CD]{%
\includegraphics[width=0.23\textwidth]{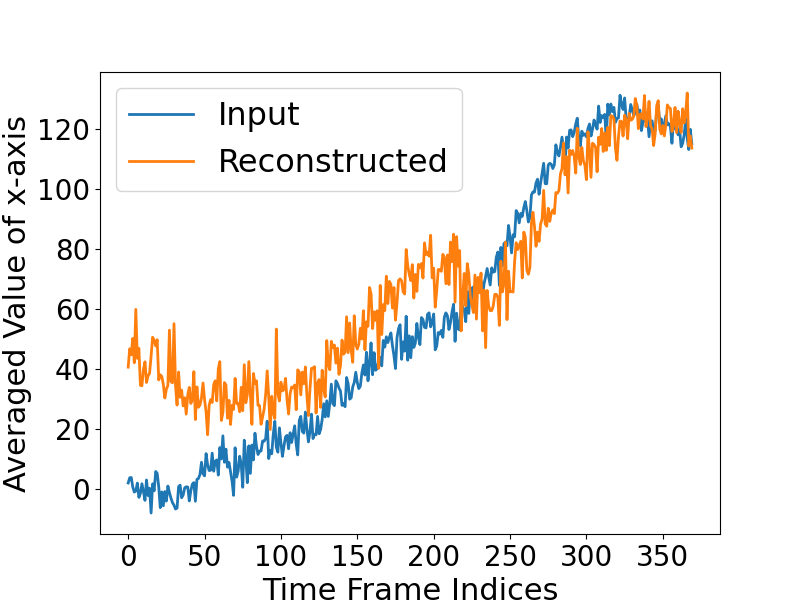}
}
\subfigure[x-axis: MCD]{%
\includegraphics[width=0.23\textwidth]{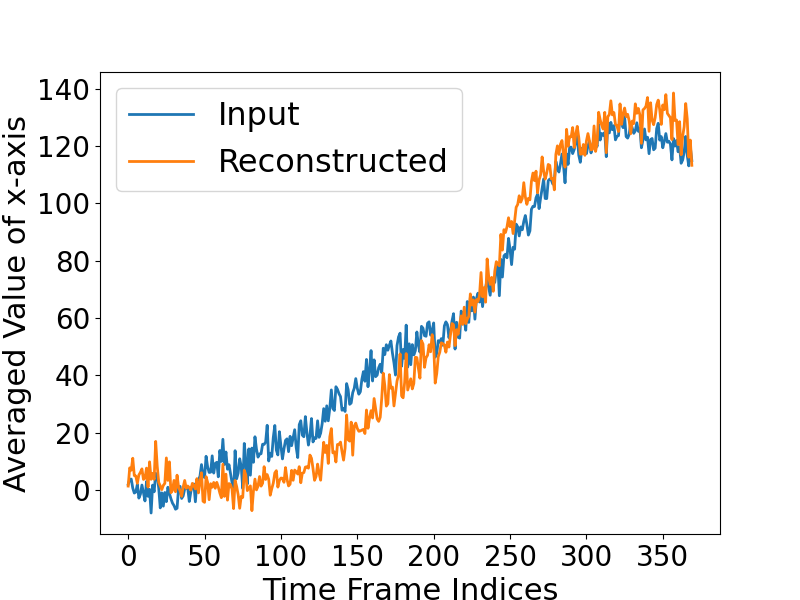}
} 
\subfigure[y-axis: CD]{%
\includegraphics[width=0.23\textwidth]{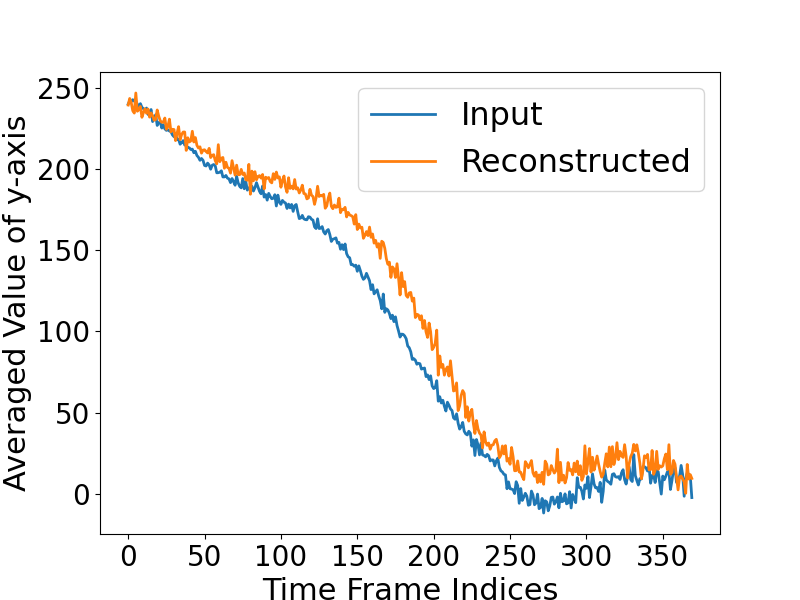}
}
\subfigure[y-axis: MCD]{%
\includegraphics[width=0.23\textwidth]{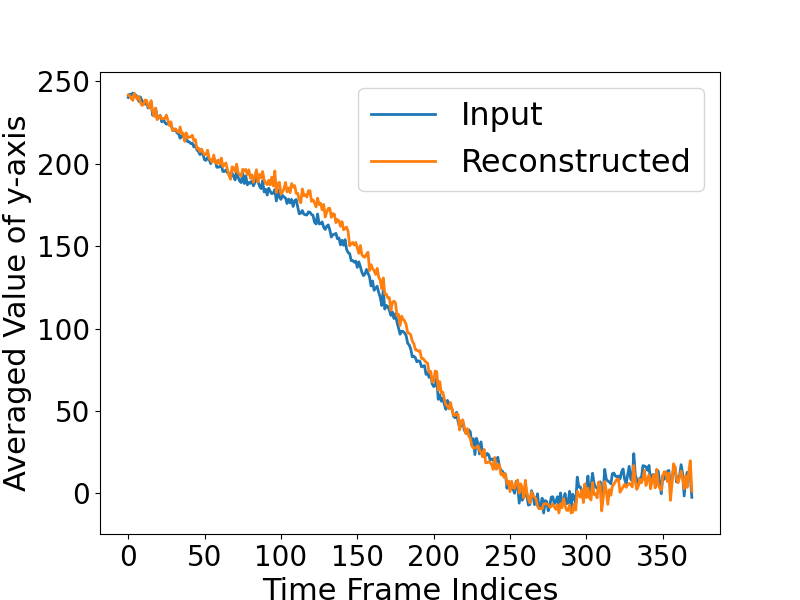}
}
}
\caption{Average values of the coordinates along x- and y-axis of the input point cloud (blue) and the corresponding reconstruction of a FoldingNet (orange) that was trained either with CD and MCD. Upon changes of the center of gravity over time, the output point cloud of MCD is more consistent with the input. This indicates that a FoldingNet trained with MCD outperforms CD in the reconstruction quality.}\label{fig:mcd}
\end{figure}

\begin{figure}[htb!]
{%
\centering
{
\subfigure[Time frame index = 0]{%
\includegraphics[width=0.45\textwidth]{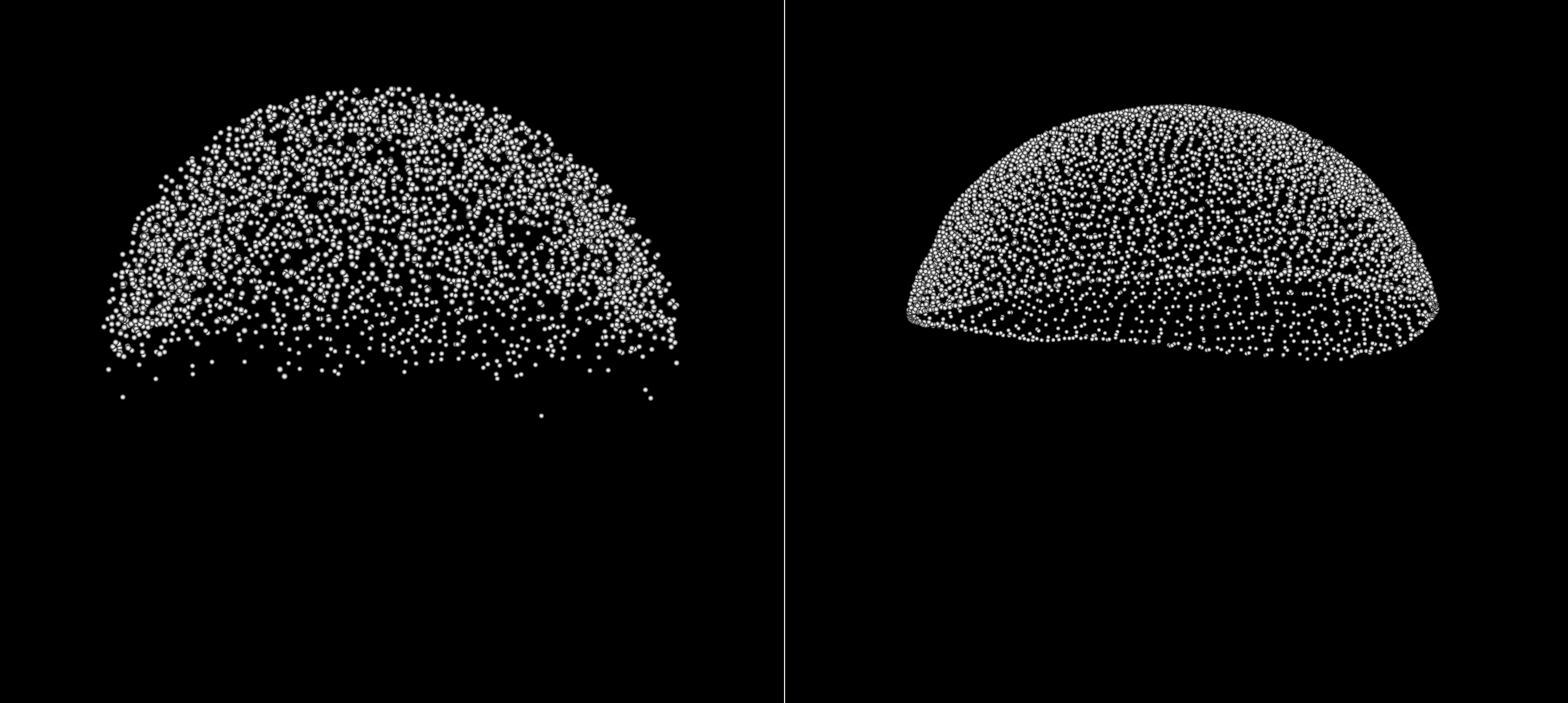}
}
\subfigure[Time frame index = 50]{%
\includegraphics[width=0.45\textwidth]{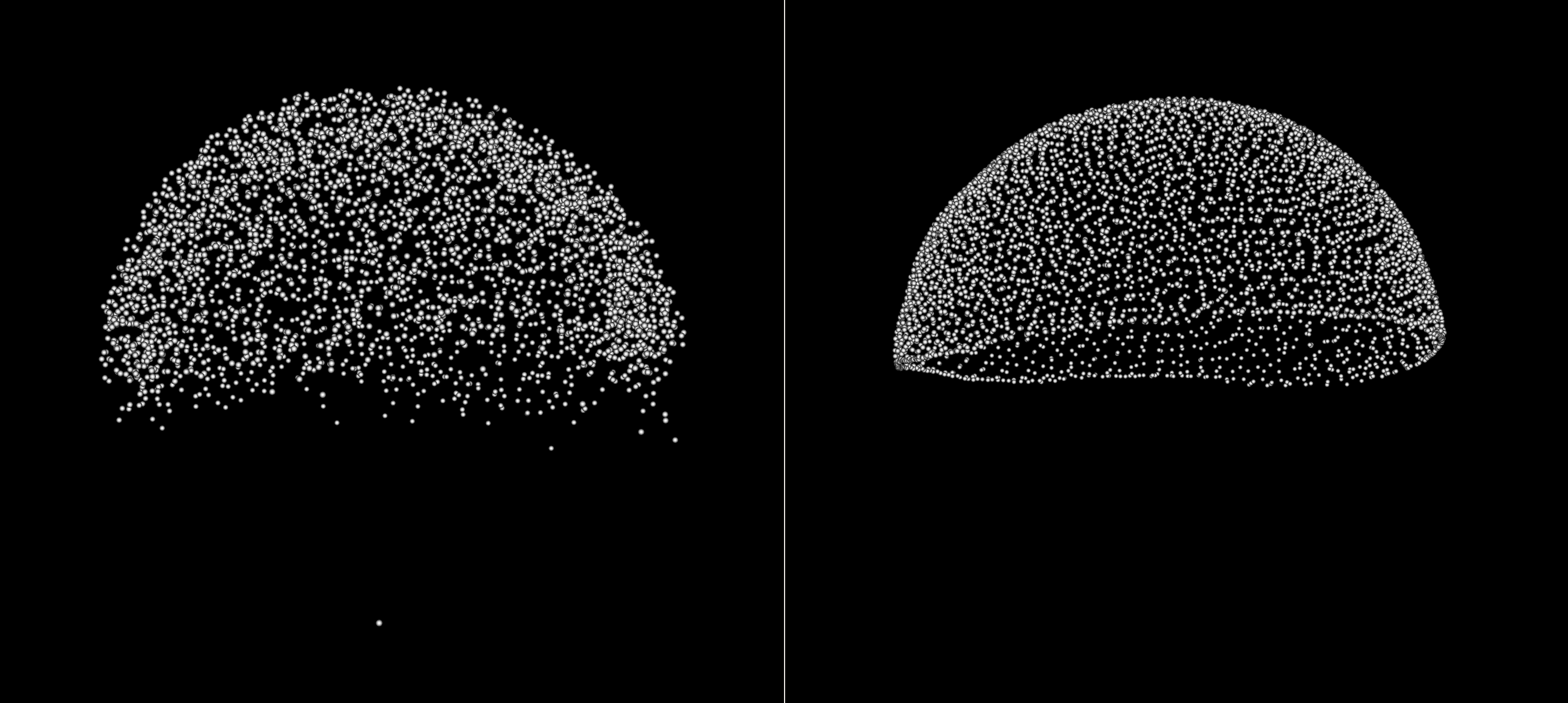}
}\\
\subfigure[Time frame index = 150]{%
\includegraphics[width=0.45\textwidth]{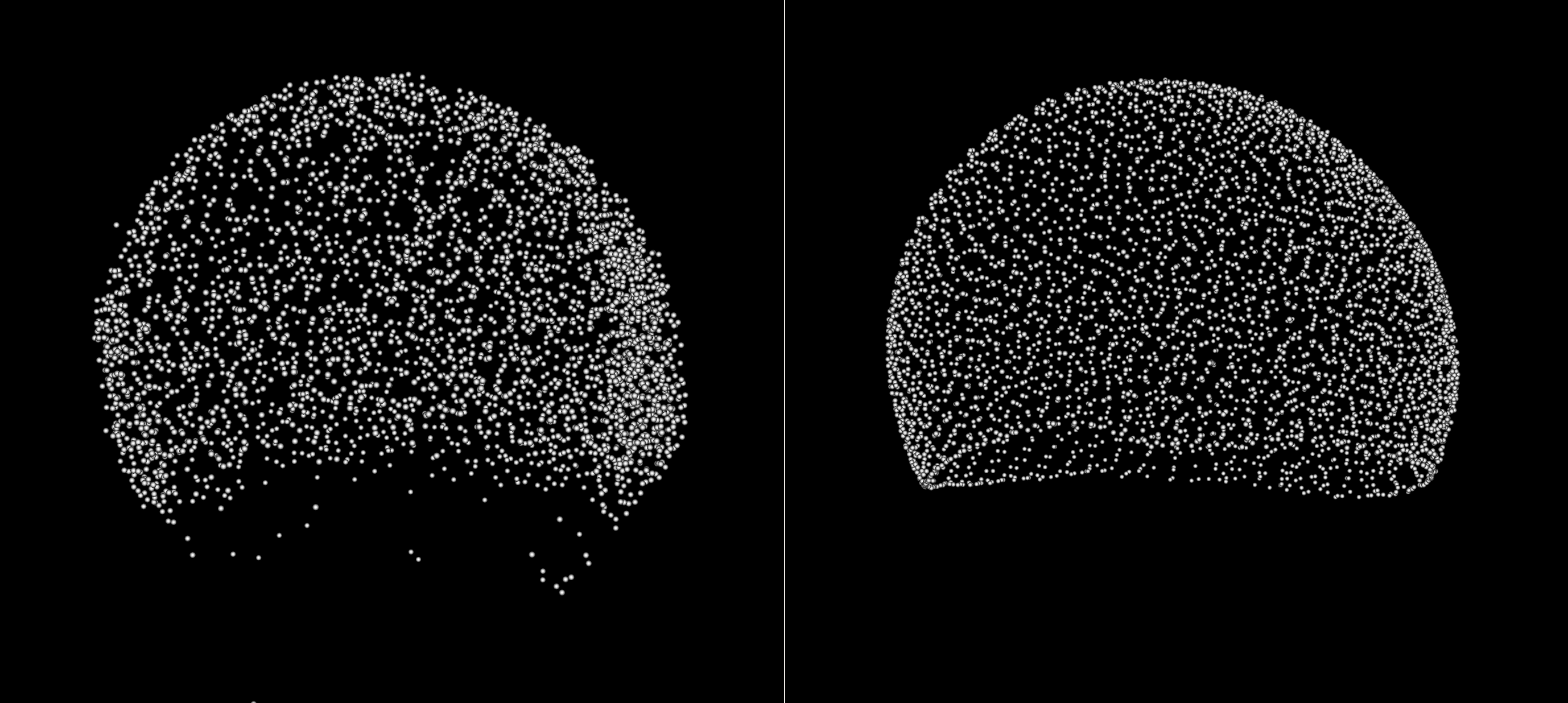}
}
\subfigure[Time frame index = 250]{%
\includegraphics[width=0.45\textwidth]{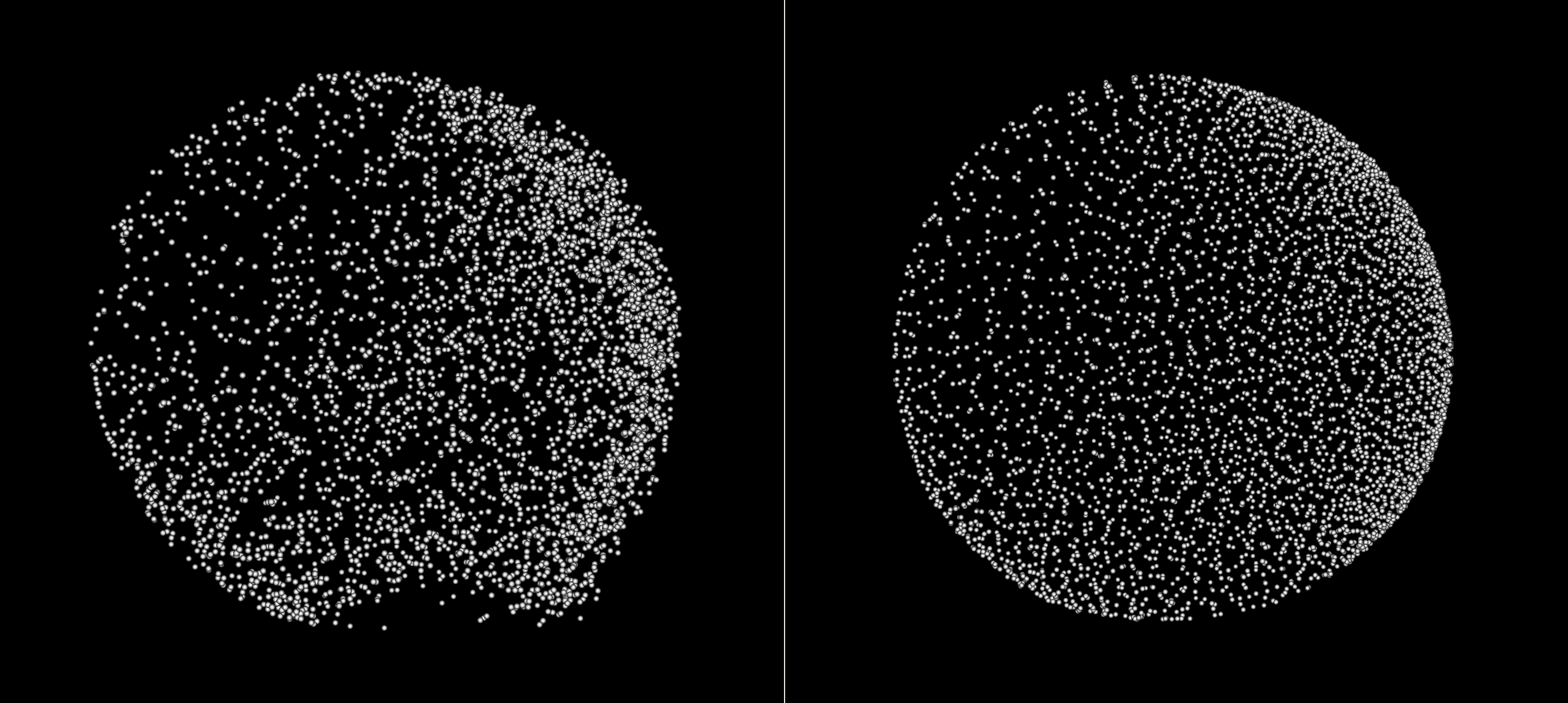}
}\\
\hspace{0.35cm}
\subfigure[Time frame index = 300]{%
\includegraphics[width=0.45\textwidth]{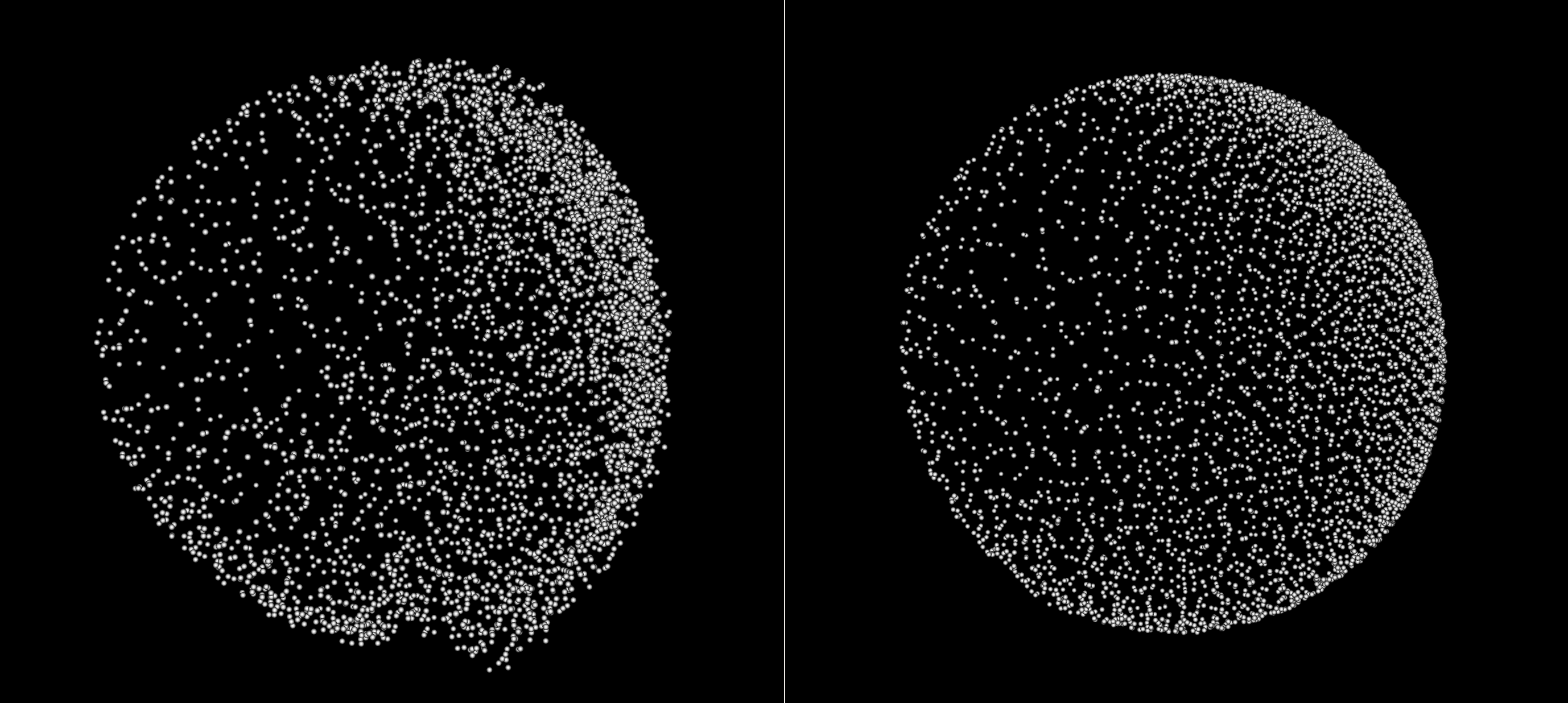}
}
\subfigure[Time frame index = 350]{%
\includegraphics[width=0.45\textwidth]{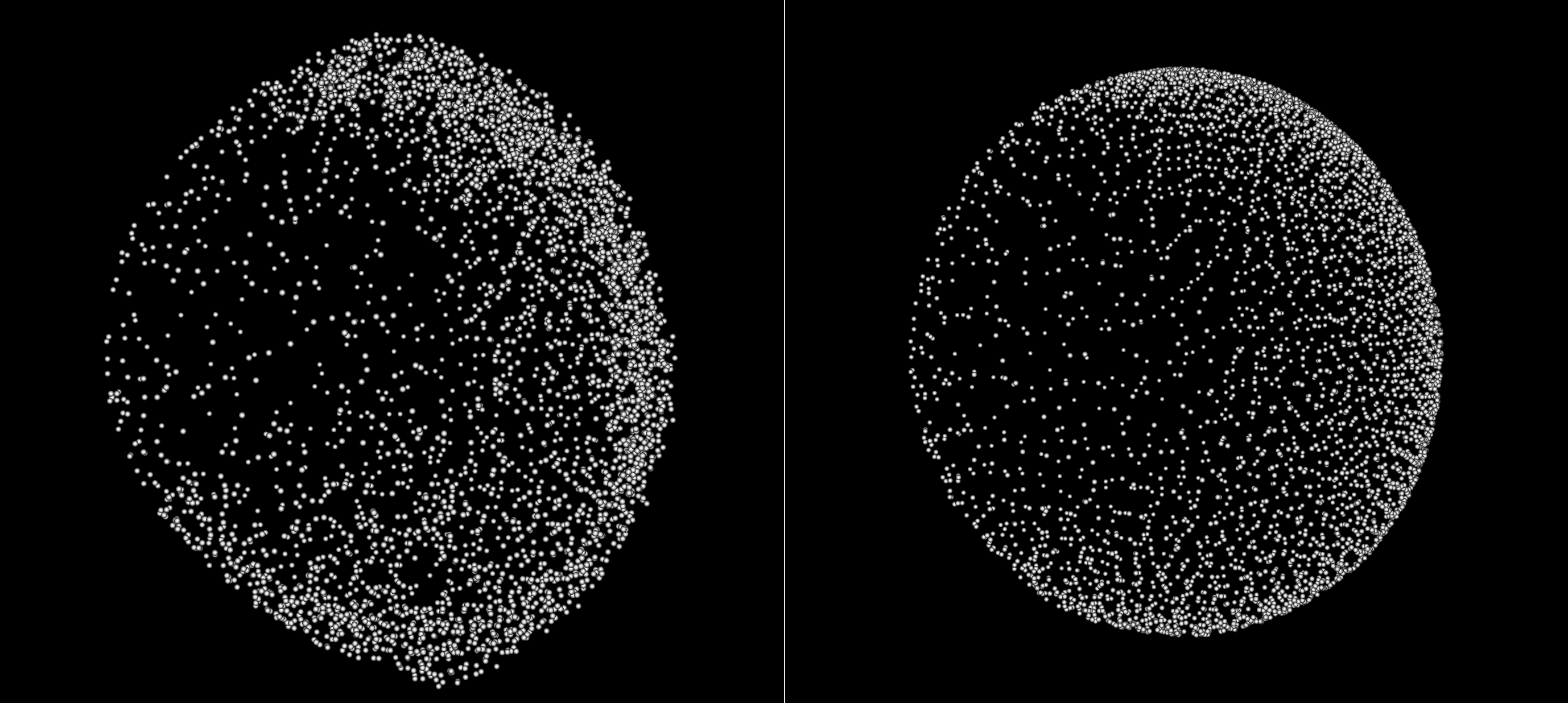}
}
}
}
\caption{Continuation of \cref{fig:draw_rec_main} in the main text.}
\label{fig:draw_rec}
\end{figure}

\begin{figure}[htb!]
{
\subfigure[Upper boundary]{%
\label{fig:lower}
\includegraphics[width=0.32\textwidth]{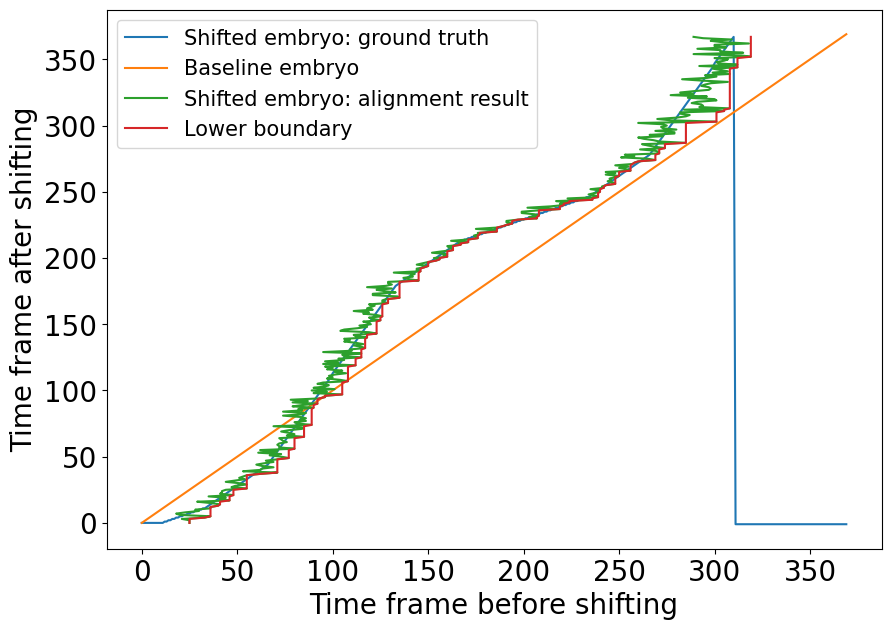}
}
\subfigure[Lower boundary]{%
\label{fig:upper}
\includegraphics[width=0.32\textwidth]{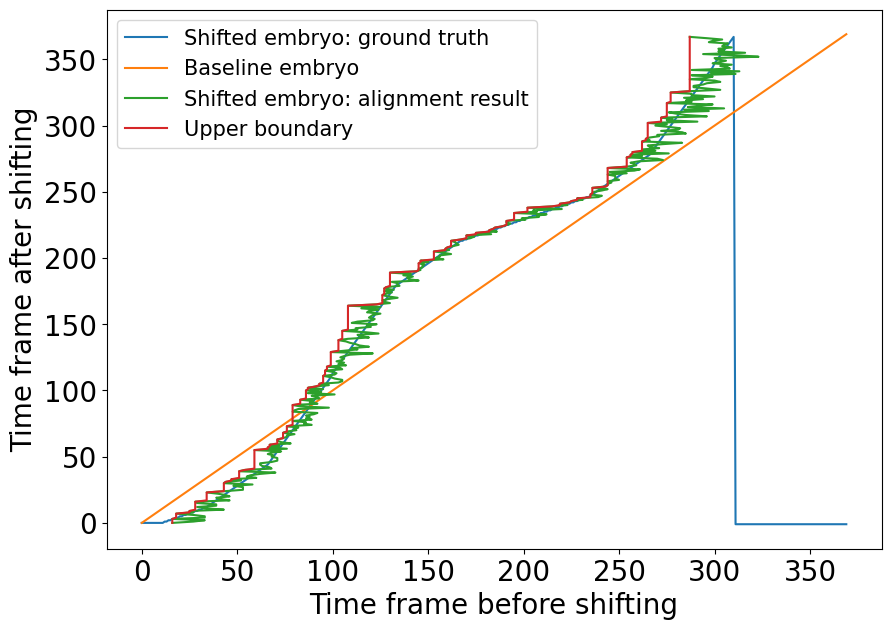}
}
\subfigure[Monot.\, inc.\, sequence]{%
\label{fig:post}
\includegraphics[width=0.32\textwidth]{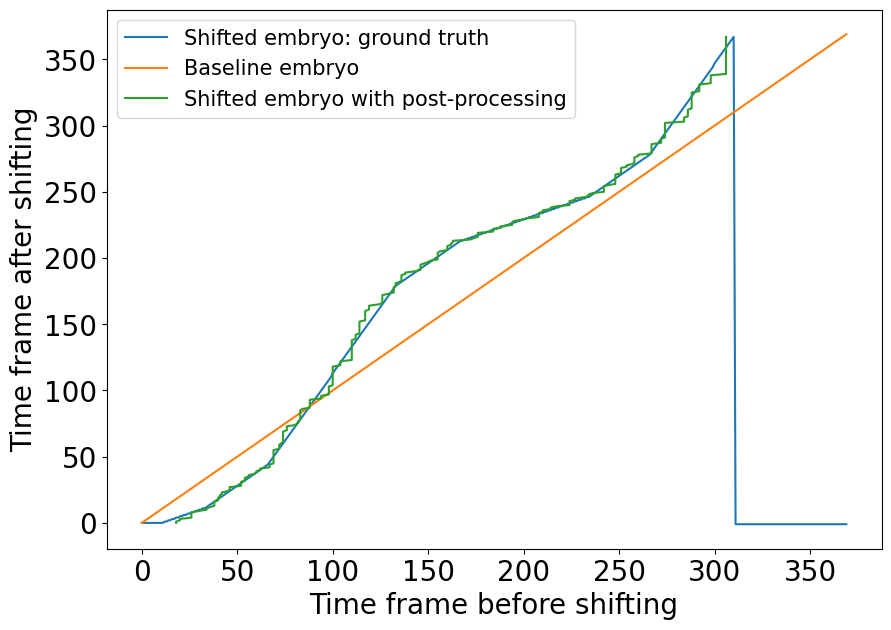}
}
}
\vspace{-0.5cm}
\caption{An example of the postprocessing. The green sequence is the alignment generated by the regression network. The lower and upper boundaries (best monotonically increasing approximation of the generated alignment below/above the green curve) are calculated and their mean value is the resulting monot.\, inc.\, sequence.}\label{fig:postprocessing}
\end{figure}

\begin{figure}[htb!]
{%
\subfigure[Cos]{%
\label{fig:cos_rot}
\includegraphics[width=0.234\textwidth]{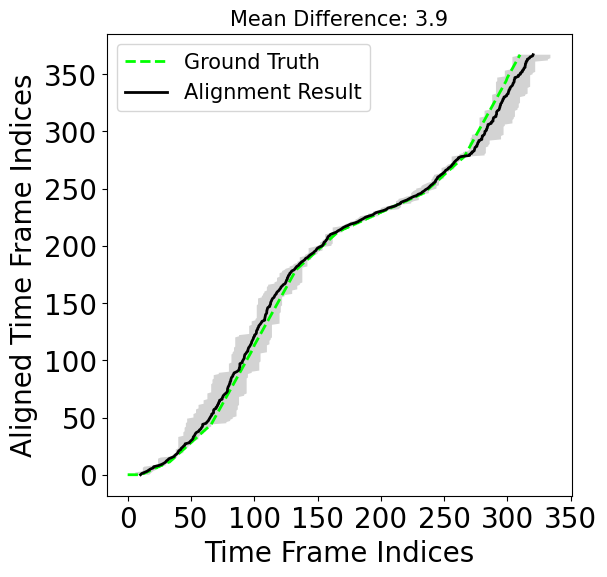}
}
\subfigure[Sin]{%
\label{fig:sin_rot}
\includegraphics[width=0.234\textwidth]{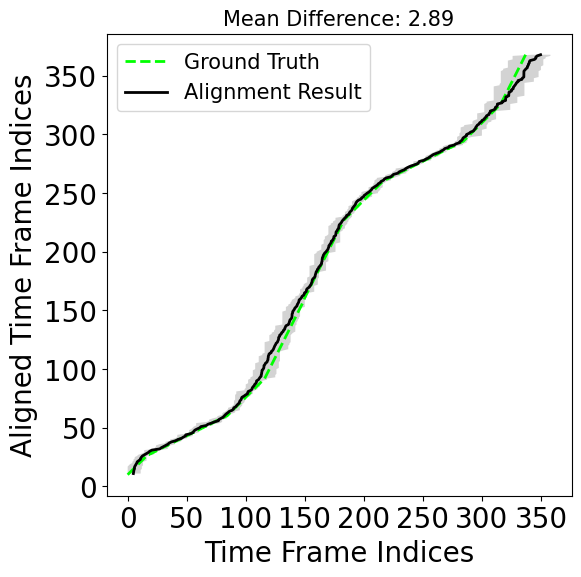}
}
\subfigure[Gaussian]{%
\label{fig:gaus_rot}
\includegraphics[width=0.234\textwidth]{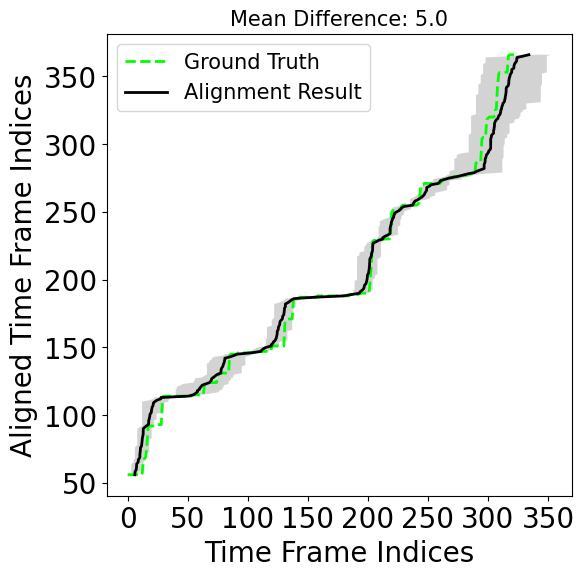}
}
\subfigure[Faster]{%
\label{fig:const_rot}
\includegraphics[width=0.234\textwidth]{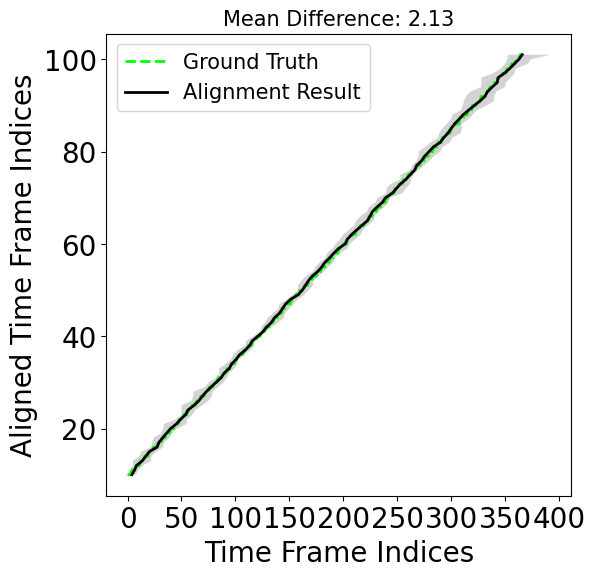}
}
}
\caption{Alignment results of the rotated embryos (MCD loss for autoencoder).}\label{fig:draw_rot}
\end{figure}

\begin{figure}[htb!]
\subfigure[Cos]{%
\label{fig:cos_rot_norm}
\includegraphics[width=0.234\textwidth]{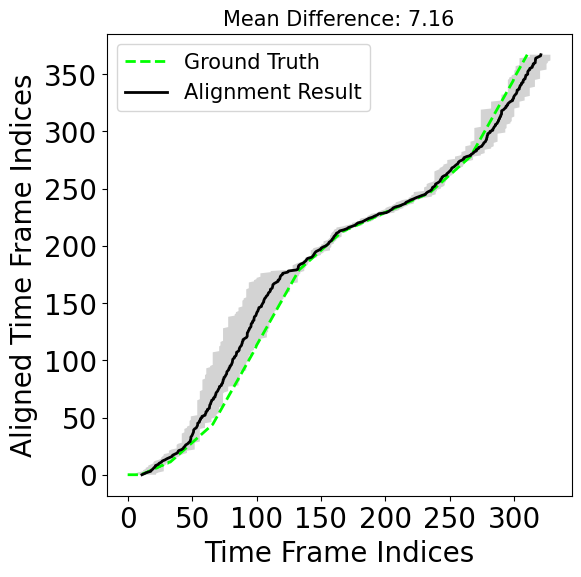}
}
\subfigure[Sin]{%
\label{fig:sin_rot_norm}
\includegraphics[width=0.234\textwidth]{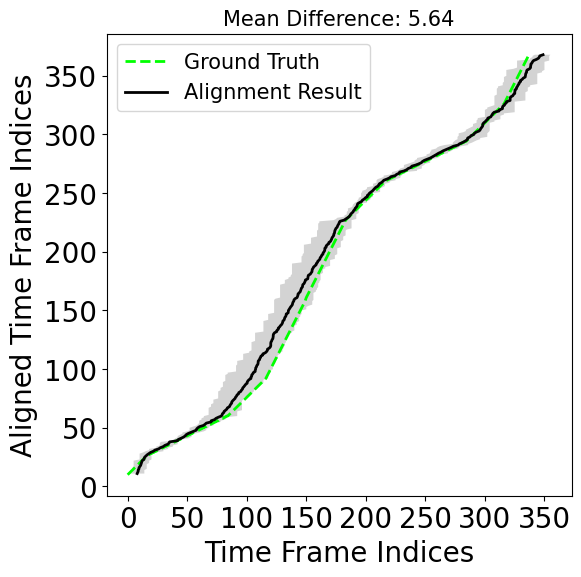}
}
\subfigure[Gaussian]{%
\label{fig:gaus_rot_norm}
\includegraphics[width=0.234\textwidth]{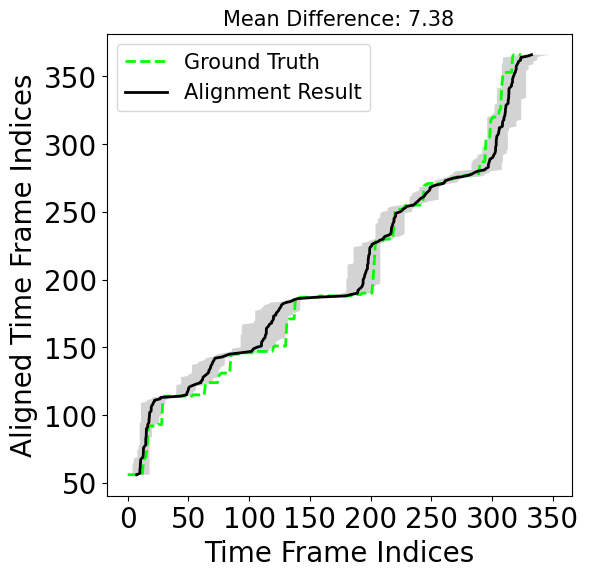}
}
\subfigure[Faster]{%
\label{fig:const_rot_norm}
\includegraphics[width=0.234\textwidth]{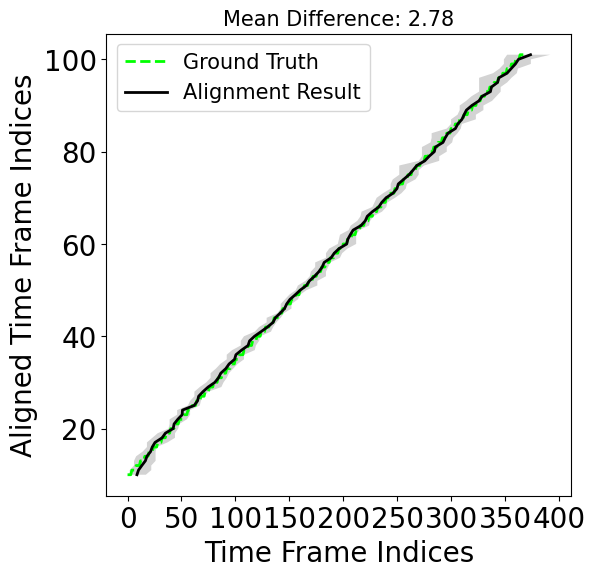}
}
\caption{Alignment results of the rotated and origin-centered embryos (MCD).}\label{fig:draw_rot_norm}
\end{figure}

\begin{figure}[htb!]
\subfigure[Cos]{
\includegraphics[width=0.225\textwidth]{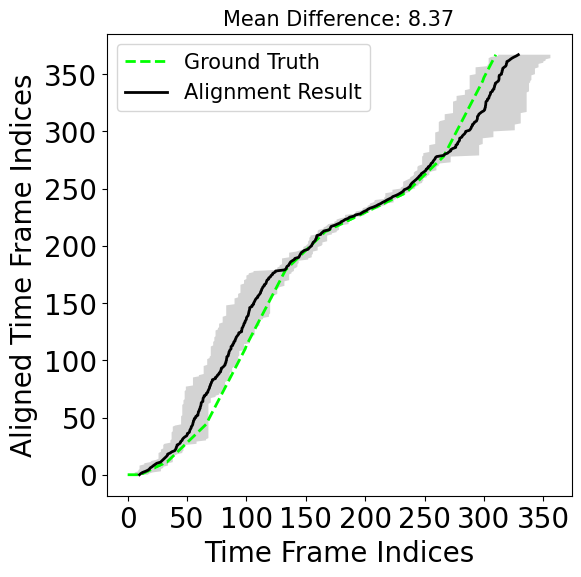}
}
\subfigure[Sin]{
\includegraphics[width=0.225\textwidth]{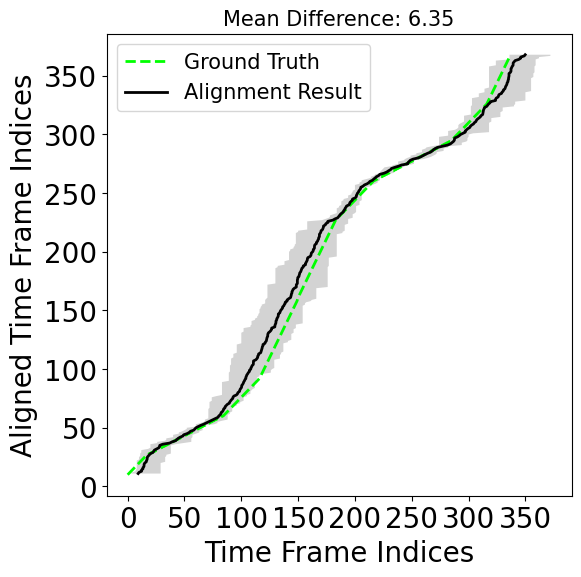}
}
\subfigure[Gaussian]{
\includegraphics[width=0.225\textwidth]{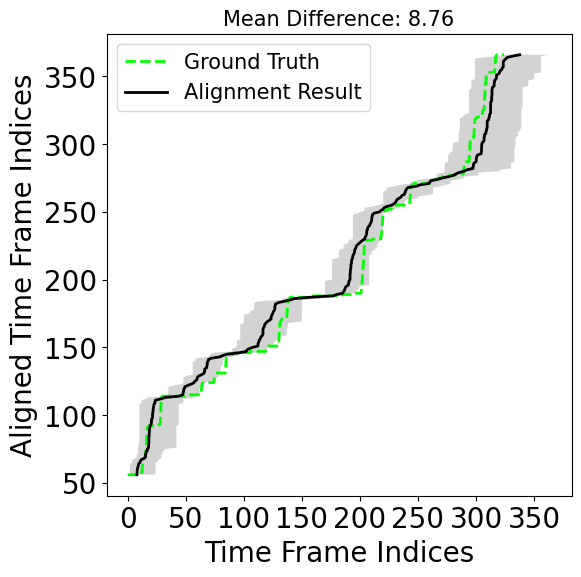}
}
\subfigure[Faster]{
\includegraphics[width=0.234\textwidth]{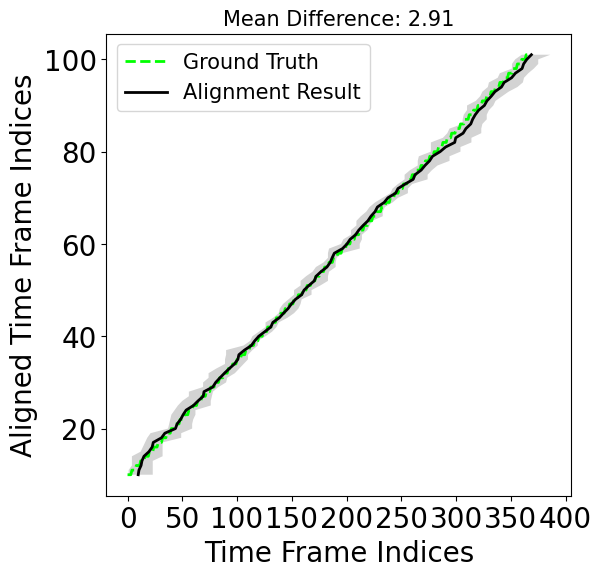}
}
\caption{Alignment results of the rotated and origin-centered embryos (feature vectors obtained from an autoencoder trained with CD loss).}\label{fig:draw_rot_norm_cd}
\end{figure}


\end{document}